\documentclass[letterpaper]{article} 
\usepackage{aaai2026}  
\usepackage{times}  
\usepackage{helvet}  
\usepackage{courier}  
\usepackage[hyphens]{url}  
\usepackage{graphicx} 
\usepackage{natbib}  
\usepackage{caption} 
\usepackage{algorithm}
\usepackage{algorithmic}
\usepackage[table]{xcolor}
\usepackage{amsmath}
\usepackage{booktabs}
\usepackage{cleveref}
\usepackage{amsfonts}
\usepackage{multirow} 
\usepackage{rotating}
\usepackage{newfloat}
\usepackage{listings}
\DeclareCaptionStyle{ruled}{labelfont=normalfont,labelsep=colon,strut=off} 
\floatstyle{ruled}
\newfloat{listing}{tb}{lst}{}
\floatname{listing}{Listing}
\title{LiNeXt: Revisiting LiDAR Completion with Efficient Non-Diffusion Architectures}
\author{
    Wenzhe He,\ 
    Xiaojun Chen,\ 
    Ruiqi Wang,\ 
    Ruihui Li\thanks{Corresponding author.},\\ 
    Huilong Pi\footnotemark[1],\
    Jiapeng Zhang,\ 
    Zhuo Tang,\ 
    Kenli Li
}
\affiliations{
    College of Computer Science and Electronic Engineering, Hunan University, China\\
    \{hewenzhe, chenxiaojun, Wrq2037825386, liruihui, phl880217, zhangjp, ztang, lkl\}@hnu.edu.cn
}

\usepackage{bibentry}
\nocopyright               

\begin{document}

\maketitle

\begin{abstract}

3D LiDAR scene completion from point clouds is a fundamental component of perception systems in autonomous vehicles. Previous methods have predominantly employed diffusion models for high‑fidelity reconstruction. However, their multi-step iterative sampling incurs significant computational overhead, limiting its real-time applicability. To address this, we propose LiNeXt: a lightweight, non‐diffusion network optimized for rapid and accurate point cloud completion. Specifically, LiNeXt first applies the Noise‑to‑Coarse (N2C) Module to denoise the input noisy point cloud in a single pass, thereby obviating the multi‑step iterative sampling of diffusion‑based methods. The Refine Module then takes the coarse point cloud and its intermediate features from the N2C Module to perform more precise refinement, further enhancing structural completeness.
Furthermore, we observe that LiDAR point clouds exhibit a distance-dependent spatial distribution, being densely sampled at proximal ranges and sparsely sampled at distal ranges. Accordingly, we propose the Distance‑aware Selected Repeat strategy to generate a more uniformly distributed noisy point cloud. On the SemanticKITTI dataset, LiNeXt achieves a 199.8 times speedup in inference, reduces Chamfer Distance by 50.7 percent, and uses only 6.1 percent of the parameters compared with LiDiff. These results demonstrate the superior efficiency and effectiveness of LiNeXt for real-time scene completion.

\end{abstract}

\section{Introduction}

Autonomous driving perception systems primarily utilize LiDAR sensors to acquire 3D point clouds of the surrounding environment, facilitating precise scene reconstruction and safe navigation. Nevertheless, the inherent sparsity of LiDAR measurements combined with frequent occlusions often leads to substantial unobserved regions in the raw point clouds. Such incompleteness hinders critical downstream tasks, including object detection~\cite{wu2022sparse, Lang_2019_CVPR, guo2025tsp3d}, pose estimation~\cite{9191119Pointvotenet}, and mapping~\cite{popovic2021volumetric}. To overcome these limitations, scene completion methods~\cite{vizzo2022makeitdense, pvd} aim to infer and reconstruct missing spatial structures, providing complete 3D representations that enhance the robustness and reliability of autonomous driving perception.

\begin{figure}[t]
    \centering
    \includegraphics[width=1.00\columnwidth]{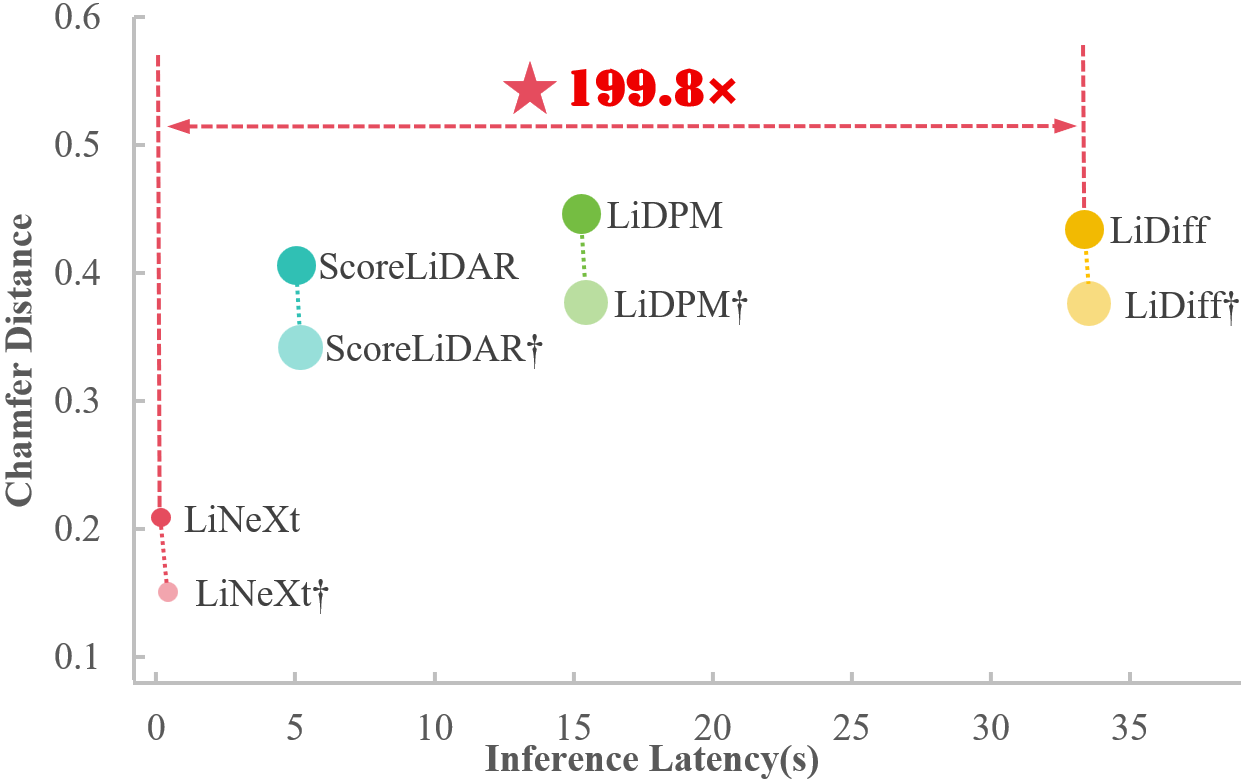} 
    \caption{LiNeXt compares reconstruction performance, inference time, and model size; the area of each marker corresponds to the number of parameters, while the symbol † indicates further refinement. For instance, compared to LiDiff~\cite{lidiff}, LiNeXt achieves 199.8$\times$ faster inference speed, a 50.7\% reduction in Chamfer Distance, and uses only 6.1\% of the parameters.
    }
    \label{fig:teaser}
\end{figure}
In scene completion research, traditional grid-based representations, such as voxel grids~\cite{Li2020aicnet, roldao2020lmscnet, zhang2023occformer} and signed distance fields~\cite{vizzo2022makeitdense, li2023lode, 10658156SurroundSDF}, have been widely employed to model 3D geometry. Voxel representations discretize scenes into regular grids, encoding geometry via occupancy or cell attributes, while SDFs, typically built on voxel grids, assign signed distances to implicitly define surfaces. However, these methods are limited by resolution trade-offs: lower resolutions fail to capture fine geometric details, whereas higher resolutions incur significant memory and computational costs. In contrast, point clouds provide a flexible, quantization‑free representation that directly encodes complex geometry and fine spatial details with high fidelity. This advantage enables the preservation of the geometric fidelity of the original scene. Building on this, recent studies, including LiDiff~\cite{lidiff}, LiDPM~\cite{martyniuk2025lidpm}, and ScoreLiDAR~\cite{ScoreLiDAR}, have adopted diffusion-based frameworks for LiDAR scene completion, leveraging the precision of point clouds to achieve promising results. Nevertheless, the iterative nature of diffusion sampling introduces substantial computational overhead, leading to slow inference times that impede real-time applications. Furthermore, the denoising objective in these models poses optimization challenges, especially when high-magnitude noise causes significant point displacement, thereby complicating accurate noise estimation and removal. By comparison, directly minimizing the Chamfer Distance (CD) provides a simpler and more effective approach to reconstructing the underlying geometry.

In this work, we forgo the use of cumbersome diffusion models and instead employ a lightweight network to reconstruct the scene. Specifically, we first introduce a Distance-aware Selected Repeat strategy, which replicates points according to their distance from the LiDAR sensor: closer points are repeated less frequently, while farther points are repeated more, resulting in a more uniform spatial distribution. We then introduce Gaussian noise to the replicated points. Subsequently, the Noise to Coarse (N2C) Module directly reconstructs the coarse scene structure, thereby avoiding the substantial time overhead incurred by the denoising process of diffusion models. The resulting output from N2C is then fed into the Refine Module to enhance structural completeness and geometric detail accuracy. Furthermore, we introduce a Cross-Point Attention (CPA) mechanism that dynamically aligns features and fuses complementary information between two input points, thereby strengthening the inference of missing structures and substantially improving completion accuracy and consistency. 
The Multi‑Scale Sparse Convolution (MSSC) module extracts point features at multiple voxel resolutions via efficient sparse convolutions and fuses them into a unified descriptor, enabling the network to capture both fine‑grained local geometry and coarse global context with minimal overhead.

Experimental results demonstrate that, as shown in \Cref{fig:teaser}, on the SemanticKITTI~\cite{semantickitti} benchmark, LiNeXt achieves a $199.8\times$ inference speedup over LiDiff, reduces the Chamfer Distance by 50.7\%, and requires only 6.1\% of LiDiff’s parameters. This combination of accelerated processing, improved reconstruction fidelity, and compact model footprint highlights LiNeXt’s efficacy for real‑time 3D scene completion in autonomous driving.

Our contributions are summarized as follows:
\begin{itemize}
  \item We develop the \textbf{Cross‑Point Attention (CPA)} and \textbf{Multi‑Scale Sparse Convolution (MSSC)} modules to form the core of the LiNeXt architecture, enabling enriched feature alignment and multi‑resolution geometric reasoning.
  
  \item We abandon traditional diffusion‑based denoising and instead employ a lightweight network to directly reconstruct complete 3D scenes.
  
  \item We demonstrate state‑of‑the‑art accuracy on both the SemanticKITTI~\cite{semantickitti} and KITTI‑360~\cite{kitti360} benchmarks, substantially outperforming existing methods.
  
  \item Experimental results demonstrate that LiNeXt achieves a $199.8\times$ inference speedup over LiDiff~\cite{lidiff} while using only 6.1\% of its parameters, highlighting its computational efficiency and practical suitability for diverse real‑world applications.

\end{itemize}

\section{Related Work}

\subsection{3D LiDAR Scene Completion}

Early completion techniques predominantly employed voxel-grid discretization or implicit signed distance field (SDF) representations. Voxel-based approaches include OccRWKV~\cite{wang2024occrwkv}, which achieves efficient semantic occupancy prediction through linear-complexity modeling and bird's-eye-view feature fusion, and OccFormer~\cite{zhang2023occformer}, leveraging dual-path transformers for 3D volume processing. Implicit SDF methods encompass Make it Dense~\cite{vizzo2022makeitdense}, a self-supervised framework completing sparse LiDAR scans to dense TSDF volumes, and SurroundSDF~\cite{10658156SurroundSDF}, performing implicit scene reconstruction via query-based SDF prediction.
While inferring missing geometry through occupancy probabilities or continuous distance fields, these methods incur substantial computational costs and remain constrained by grid resolution, which blurs fine structures. 

\subsection{Single-Object Point Cloud Completion}

PCN \cite{yuan2018pcn} pioneered learning latent shape representations for missing point generation. Subsequent works, using a coarse-to-fine\cite{xie2020grnet, pan2021variational, wen2021pmp, xiang2022snowflake, zhou2022seedformer, rong2024cra, 25_ijcai_iaet} manner, have significantly advanced robustness and detail fidelity. For example, SnowflakeNet \cite{xiang2022snowflake} leverages Snowflake Point Deconvolution (SPD) for hierarchical point upsampling, where skip-transformer layers learn point splitting patterns to recover fine-grained geometric details.
AdaPoinTr \cite{Yu2023AdaPoinTr} reformulates completion as set-to-set translation using a Transformer encoder-decoder structure.
PointAttN \cite{Wang2024PointAttN} refines local geometric dependency capture through structured self-attention mechanisms.
Most recently, GenPC \cite{Li_2025_CVPR_GenPC} leverages 3D generative priors with depth prompting and geometric fusion to achieve zero-shot completion on real-world scans. Existing single-object point cloud completion methods focus on virtual models and often fail in complex real-world scenarios.

\begin{figure*}[t!]
  \centering
  \includegraphics[width=\textwidth]{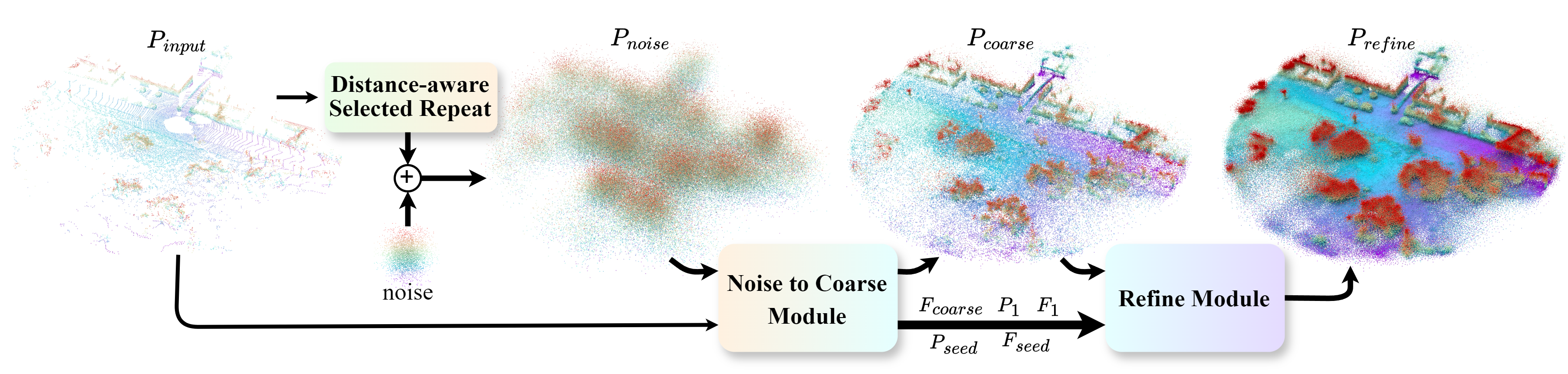}
  \caption{Overall framework of LiNeXt.}
  \label{fig:PipeLine}
\end{figure*}

\subsection{Diffusion Models in Completion}
Recently, an increasing number of research endeavors~\cite{ni2025dprecon, cao2024diffsscsemanticlidarscan, du2025superpc, zhao2025diffusiondistillationdirectpreference} have focused on utilizing diffusion modules for scene completion. For example, PVD~\cite{pvd} unifies shape generation and completion via a probabilistic point-voxel diffusion model, but it demonstrates limited effectiveness in outdoor scene completion. LiDiff~\cite{lidiff} and its variants adopt a locally guided diffusion process within the DDPM framework, yielding marked improvements in outdoor reconstruction fidelity. LiDPM~\cite{martyniuk2025lidpm} further rethinks this paradigm, proving vanilla DDPMs with proper initialization achieve superior scene completion without local diffusion approximations. 
Critically, LiDiff and LiDPM exhibit high inference latency. 
ScoreLiDAR~\cite{ScoreLiDAR} employs knowledge distillation to accelerate diffusion-based sampling, making completion 5$\times$ faster while maintaining competitive completion quality. Despite their training stability and high output fidelity, diffusion-based methods typically require slow sampling processes and complex network architectures, making them impractical for large-scale real-time perception systems.

\section{Method}

As illustrated in Figure~\ref{fig:PipeLine}, the incomplete LiDAR point cloud \(P_{input}\) is duplicated in the Distance-aware Selected Repeat strategy (less for near, more for far) and corrupted with Gaussian noise to yield \(P_{noise}\). The Noise‐to‐Coarse Module denoises \(P_{noise}\) under supervision of \(P_{input}\), producing \(P_{coarse}\). The Refine Module then refines this output into the final high‐quality point cloud \(P_{refine}\).

\subsection{Distance-aware Selected Repeat}

Existing diffusion-based~\cite{lidiff,ScoreLiDAR} methods replicate the input point cloud \(\,P_{input}\) uniformly, resulting in a noisy cloud \(P_{noise}\) that oversamples near‑field points while undersampling far‑field points. To address this imbalance, we propose a Distance‑Aware Selected Repeat (DSR) strategy, which adjusts duplication factors based on each point’s Euclidean distance from the origin, thereby yielding a more uniformly distributed \(P_{noise}\).
Formally, let \(P_{input}=\{p_i\}_{i=1}^N\), and compute
\begin{equation}
d_i = \lVert p_i \rVert,\quad i = 1, \dots, N.
\end{equation}
Sort the points by \(d_i\) in ascending order to obtain \(\{p_{(1)},\dots,p_{(N)}\}\), then partition them into four equal-sized groups:
\begin{equation}
G_k = \bigl\{\,p_{((k-1)\frac{N}{4} + 1)}, \dots, p_{(k\frac{N}{4})}\bigr\},\quad k = 1,2,3,4.
\end{equation}

Assign duplication counts of \(\{r_1=5, r_2=8, r_3=12, r_4=15\}\) to groups \(\{G_1, G_2, G_3, G_4\}\), respectively, forming the replicated set \(P_{rep}\). Finally, add independent Gaussian noise to each point in \(P_{rep}\), yielding the balanced noisy point cloud \(P_{noise}\). This “fewer for near, more for far” replication ensures uniform coverage across distances and provides richer, evenly distributed samples for subsequent Noise-to-Coarse (N2C) modules.

\subsection{Multi-Scale Sparse Convolution (MSSC)}
The MSSC module hierarchically aggregates point‑cloud features via parallel spatially sparse convolutions applied over multiple voxel resolutions \(g_k \in \mathcal{G} = \{0.01 \times 2^{i-1} \mid i = 1,2,\dots,N_{vox}\},\) where \(N_{vox}\) denotes the total number of voxel scales. Given input coordinates \(P\in\mathbb{R}^{N\times3}\), initial point‑wise features are computed as \(X = \mathrm{MLP}_{\mathrm{init}}(P).\) For each resolution \(g_k\in\mathcal{G}\), the point cloud is voxelized and embedded:
\begin{equation}
\hat{P}_k = \bigl\lfloor P / g_k \bigr\rfloor,\qquad
F_k = \mathrm{MLP}_k(X),
\end{equation}
where \(\lfloor\cdot\rfloor\) denotes element-wise floor division. Sparse tensors \(\mathcal{T}_k\) are then constructed using the grid indices \(\hat{P}_k\) and features \(F_k\). Each \(\mathcal{T}_k\) undergoes dual residual sparse convolutions:
\begin{equation}
\begin{aligned}
\mathcal{T}'_k &= \mathrm{spconv}_{k,1}(\mathcal{T}_k) + \mathcal{T}_k, \\
\mathcal{T}''_k &= \mathrm{spconv}_{k,2}(\mathcal{T}'_k) + \mathcal{T}'_k,
\end{aligned}
\end{equation}
We denote concatenation across scales by the double‑bar operator. The multi‑scale output for each resolution is defined as \(O_k = \mathcal{T}''_k + F_k,\)
and these outputs are concatenated and projected to a unified descriptor: \(F = \mathrm{MLP}_{end}(\mathrm{CONCAT}_{k=1}^{|\mathcal{G}|}(O_k)).\) This design captures spatial hierarchies through optimized sparse 3D convolutions while preserving geometric fidelity via residual connections.

\subsection{Cross-Point Attention Module}

\begin{figure}[t]
\centering
\includegraphics[width=1.00\columnwidth]{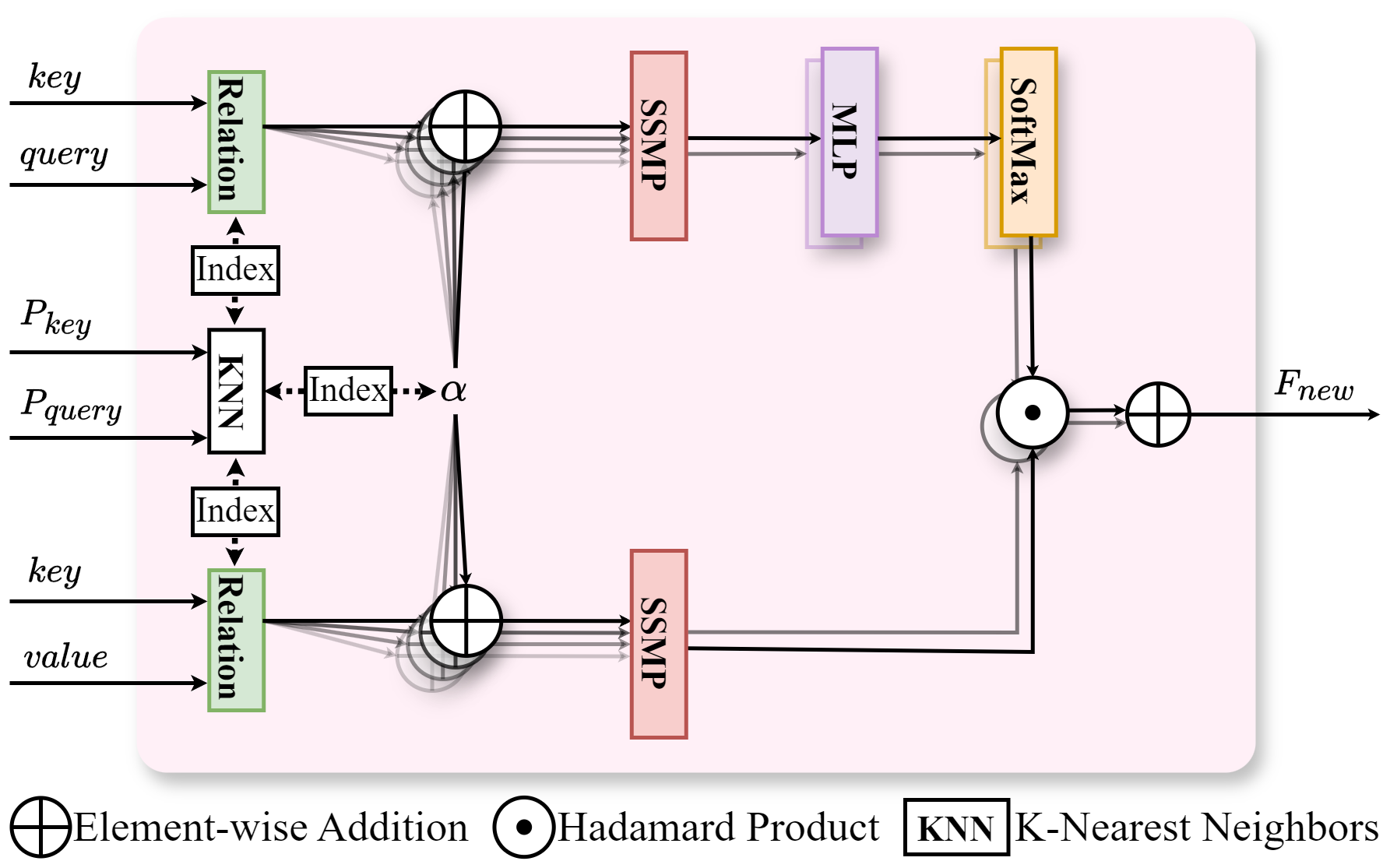} 
\caption{The structure of the Cross-Point Attention (CPA) module.}
\label{fig:CPA}
\end{figure}

The Cross-Point Attention (CPA) Module enables robust feature fusion between the global scene and localized part representations by explicitly encoding spatial relationships and leveraging attention for correspondence. As illustrated in \Cref{fig:CPA}, CPA operates on primary point cloud coordinates $P_{key}\in\mathbb{R}^{N\times3}$ with $key$, and part coordinates $P_{query}\in\mathbb{R}^{M\times3}$ with $query$ and $value$.

First, local correspondences are identified by performing a k-nearest neighbors search between the two coordinate sets, yielding index maps for grouping:
\begin{equation}
idx=\mathrm{KNN}(P_{query},P_{key},k).
\end{equation}
Using these indices, the relative displacement of each primary point with respect to its neighbors in the part set is computed to obtain spatial embedding:
\begin{equation}
\alpha=\mathrm{MLP}_{pos}(P_{key}-\mathcal{G}(P_{key},idx)).
\end{equation}
This spatial embedding $\alpha$ augments the geometry-aware feature differences:
\begin{equation}
\begin{split}
Q_{rel} &= query - \mathcal{G}(key,idx) + \alpha, \\
V_{rel} &= value - \mathcal{G}(key,idx) + \alpha.
\end{split}
\end{equation}

\begin{figure}[t]
\centering
\includegraphics[width=1.00\columnwidth]{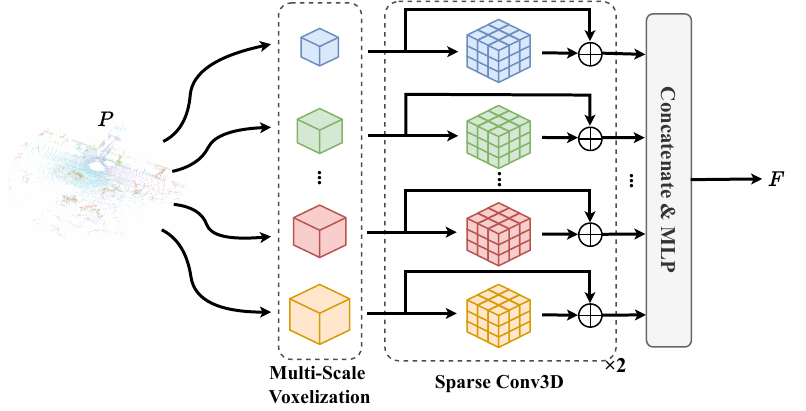} 
\caption{The structure of the Multi-Scale Sparse Convolution (MSSC) module.}
\label{fig:MSSC}
\end{figure}

Before Serial‑Segment Max‑Pooling (SSMP) processing, the input point cloud is serialized into a linear sequence using randomized spatial orderings, specifically Z-order~\cite{1966z-order} and Hilbert-order~\cite{hilbert1935stetige}, to preserve spatial locality. Next, for each point in this sequence, its local neighborhood is retrieved via k‑nearest neighbors (KNN) while preserving the original sequence order within each neighborhood, producing relational feature $Q_{rel},\,V_{rel}\;\in\;\mathbb{R}^{N\times C\times K},$
where \(N\) is the number of points, \(C\) is the feature dimension, and \(K\) is the number of neighbors. These tensors are partitioned along the dimension of the neighborhood into $\hat{K}$ segments of size $K/\hat{K}$, resulting in reshaped tensors of dimension $\mathbb{R}^{N \times C \times \hat{K} \times (K/\hat{K})}$. SSMP is then applied to compress each segment to its maximally activated feature:
\begin{equation}
\hat{Q}_{rel} = \mathrm{SSMP}(Q_{rel}), \quad 
\hat{V}_{rel} = \mathrm{SSMP}(V_{rel})
\end{equation}
where SSMP operates along the partitioned dimension $K/\hat{K}$, producing output tensors $\hat{Q}_{rel}, \hat{V}_{rel} \in \mathbb{R}^{N \times C \times \hat{K}}$.

This compression achieves critical dimensionality reduction for computational efficiency while preserving discriminative patterns, concurrently enhancing robustness to local perturbations through emphasis on dominant spatial-relationship signatures within each region.

$\hat{Q}_{rel}$ are then combined through a multi-layer perceptron and softmax to produce normalized attention weights:
\begin{equation}
\mathcal{A}=\mathrm{SoftMax}\left(\mathrm{MLP}_{attn}(\hat{Q}_{rel})\right).
\end{equation}

Finally, the attention weights modulate the value embeddings to aggregate context‑aware features, which are then fused with the original representation via a residual connection:
\begin{equation}
F_{new} = {value} + \sum_{j=1}^{\hat{K}} A_{j} \odot \hat{V}_{j},
\label{eq:cpa_fusion}
\end{equation}
where $j$ represents the index of the local segment and $\hat{K}$ denotes the total number of segments.

Through this ordered sequence of embedding, grouping, pooling, and attention, the CPA Module preserves geometric coherence while propagating complementary features across point sets, substantially enhancing the network’s ability to infer and complete missing structures.

\subsection{Noise to Coarse (N2C) Module}
\label{sec:n2c}
\begin{figure*}[h]
  \centering
  \includegraphics[width=\textwidth]{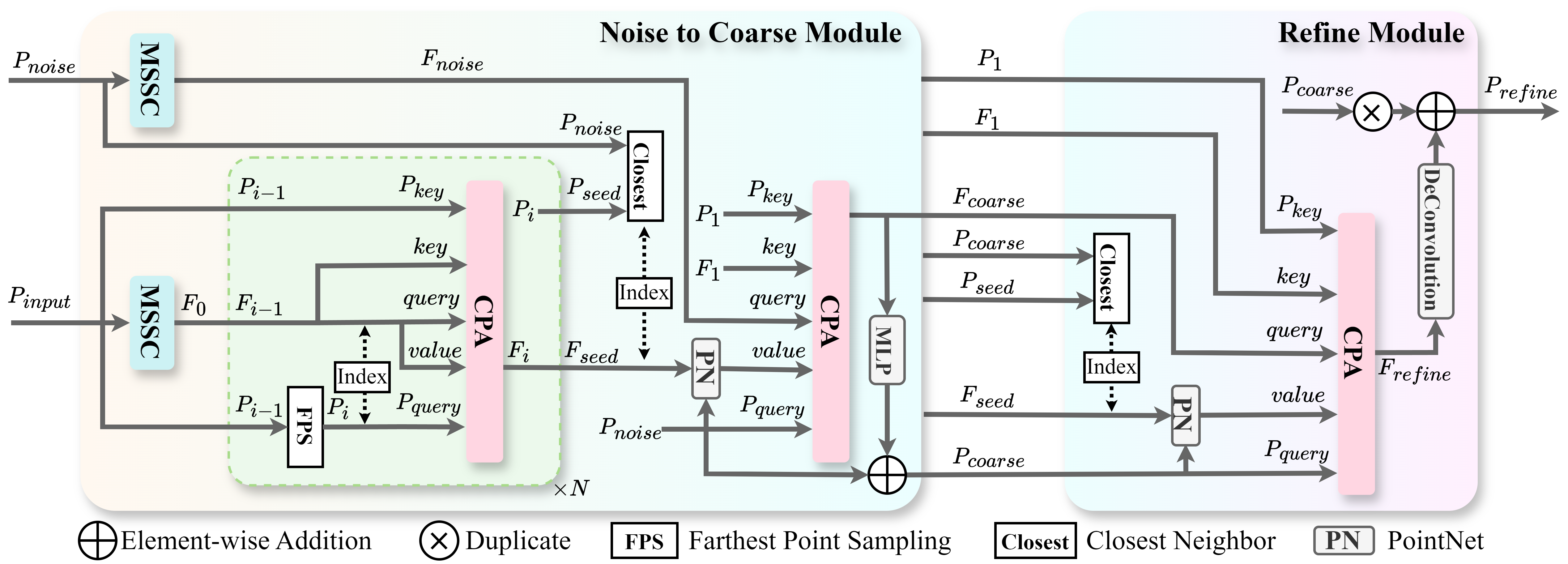}
  \caption{Detailed architectures of the Noise to Coarse Module and the Refine Module.}
  \label{fig:N2C_Refine}
\end{figure*}

As illustrated in Figure~\ref{fig:N2C_Refine}, the N2C module generates a coarse denoised point cloud by hierarchically distilling structural priors from the input distribution. Its operational pipeline comprises three core stages.

\textbf{Initial Feature Extraction:}
Multi-Scale Sparse Convolution (MSSC) extracts preliminary spatial features from the input point cloud $P_{input}$ and noisy observation $P_{noise}$, yielding enhanced feature sets $F_0$ and $F_{noise}$, respectively:
\begin{equation}
F_0 = \text{MSSC}(P_{input}), \quad F_{noise} = \text{MSSC}(P_{noise}).
\end{equation}

\textbf{Hierarchical Seed Generation:}
An $N$-stage downsampling process iteratively condenses $P_{input}$ into regional seed points $P_{seed}$ with global features $F_{seed}$. At stage $i$:
\begin{align}
\bigl(P_i,\hat{F}_{i-1}\bigr)
&=
\begin{cases}
(P_{input},F_0) & i=1,\\[2pt]
\bigl(\text{FPS}(P_{i-1}),\,F_{i-1}[index_{FPS}]\bigr) & i\ge 2.
\end{cases}
\end{align}
Here $\text{FPS}(\cdot)$ denotes farthest point sampling, and $[index_{FPS}]$ indexes the features of the sampled points. The initial stage ($i=1$) skips downsampling to allow deeper feature extraction.

Subsequently, Cross-Point Attention (CPA) processes the geometric relationships between hierarchical point sets: for each stage $i$, $P_i$ serves as the query points with features $\hat{F}_{i-1}$, while $P_{i-1}$ and its features $F_{i-1}$ provide the structural key and value. This attention mechanism outputs enhanced features $F_i$.
The final outputs \(P_{seed} = P_N\) and \(F_{seed} = F_N\) encapsulate the global features of local seed regions distilled by the iterative attention refinement process.

\textbf{Coarse Reconstruction}
Given a noisy point set $P_{noise}$ with features $F_{noise}$, we first retrieve the nearest seed points $\hat{P}_{seed}$ and their associated features $\hat{F}_{seed}$. A lightweight PointNet~\cite{qi2016pointnet} then fuses noisy coordinates with the aggregated seed features, producing relation-aware features $value$.

In the last CPA module, we treat \(F_{noise}\), \(F_{1}\), and a relation‑aware $value$ as the query, key, and value, respectively. This block jointly regresses the coarse coordinates \(P_{coarse}\) and features \(F_{coarse}\), then forwards the intermediate tensors \(\{P_{1}, F_{1}, P_{seed}, F_{seed}\}\) to the Refine Module for further detail recovery.

\begin{figure*}[t!]
  \centering
  \includegraphics[width=0.88\textwidth]{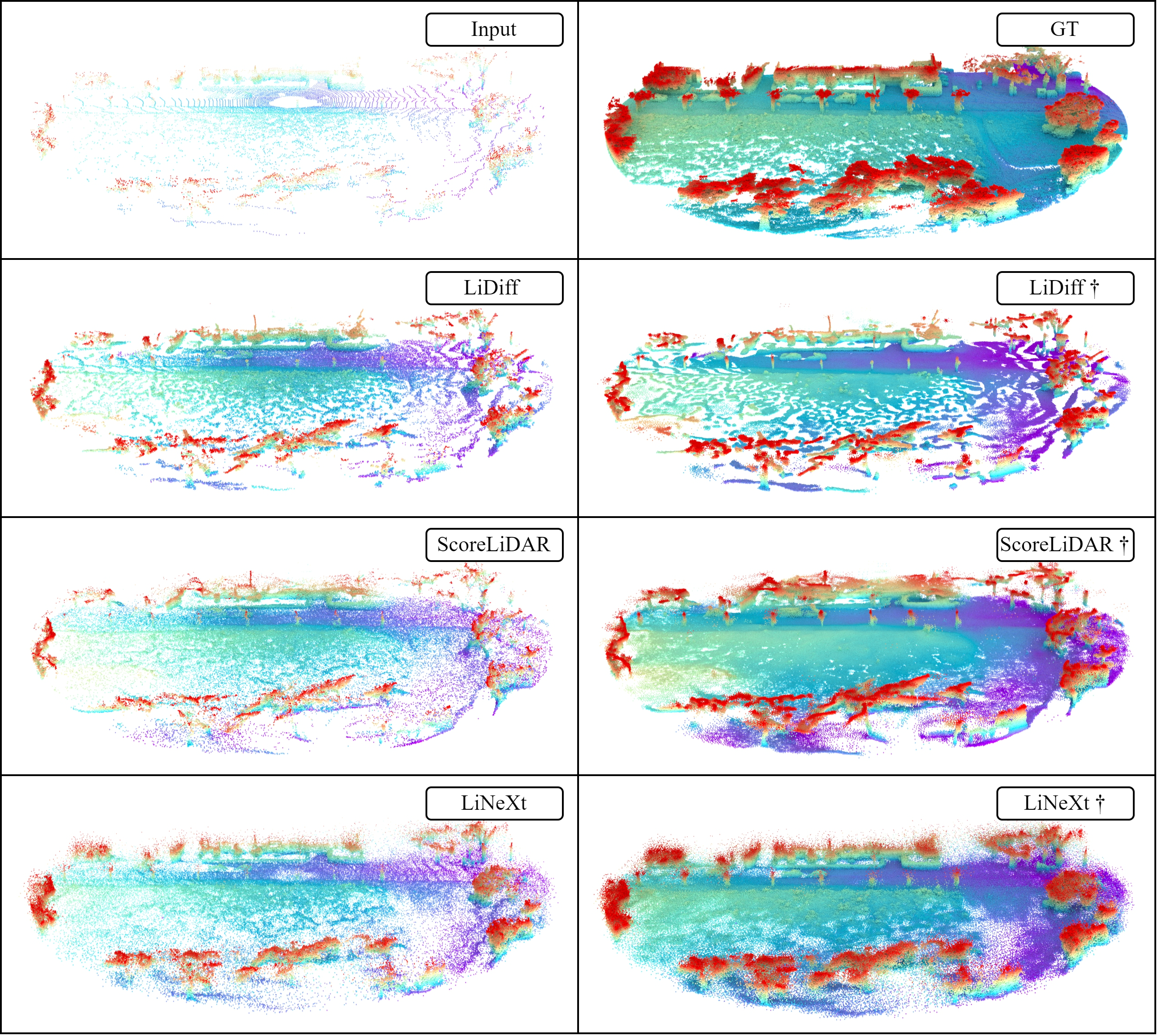}
  \caption{Visualization comparison of our method against LiDiff and ScoreLiDAR on the SemanticKITTI dataset. \textsuperscript{†} indicates additional refinement.}
  \label{fig:Compare}
\end{figure*}

\subsection{Refine Module}
The Refine Module, as shown in Figure~\ref{fig:N2C_Refine}, enhances the coarse output $P_{coarse}$ from the N2C module. For each point in $P_{coarse}$, we retrieve regional features $\hat{F}_{seed}$ from its nearest neighbor in the seed set $P_{seed}$. These seed features are combined with $P_{coarse}$ coordinates through a lightweight PointNet~\cite{qi2016pointnet} module to generate relationship-aware $value$. The CPA mechanism then processes $P_{coarse}$ as query points with features $F_{coarse}$, using $P_1$ (from the N2C's first downsampling stage) as structural keys with features $F_1$, and $value$ as geometric-relationship values. The CPA outputs refined features \(F_{refine}\), which are then passed through a deconvolution~\cite{xiang2022snowflake} module to generate the upsampled point cloud \(P_{refine}\), effectively combining local details with global context to correct residual noise while recovering geometrically consistent structures.

\subsection{Training Loss}

The ground‑truth point cloud is downsampled via voxel grid to $180{,}000$ points to ensure spatial uniformity and reduce computational cost. For each output point set $P\in\{P_{\mathrm{coarse}},\,P_{\mathrm{refine}}\}$, we compute the Chamfer Distance as

\begin{equation}
L_{\mathrm{CD}}(P,\widehat{P})
= \frac{1}{|P|}\sum_{x\in P}\min_{y\in\widehat{P}}\|x - y\|_2^2
+ \frac{1}{|\widehat{P}|}\sum_{y\in\widehat{P}}\min_{x\in P}\|y - x\|_2^2,
\end{equation}

where $\widehat{P}$ is the downsampled ground truth and $\|\cdot\|_2$ denotes the Euclidean norm. Each stage is trained independently using this loss.

\begin{table*}[h]
    \centering
    \begin{tabular}{@{}c|lcccccc@{}}
    \toprule
    &Method&CD$\downarrow$&JSD 3D$\downarrow$&JSD BEV$\downarrow$&Vox.IoU(0.5\,m)$\uparrow$&Vox.IoU(0.2\,m)$\uparrow$&Vox.IoU(0.1\,m)$\uparrow$\\
    \midrule
    \multirow{13}[0]{*}{\begin{sideways}\rotatebox{0}{SemanticKITTI}\end{sideways}} 
    &LMSCNnet&0.641&--&0.431&30.83&12.09&3.65\\
    &LODE&1.029&--&0.451&33.81&16.39&5.00\\
    &MID&0.503&--&0.470&31.58&22.72&13.14\\
    &PVD&1.256&--&0.498&15.91&3.97&0.60\\
    &LiDiff&0.434&0.564&0.444&31.47&16.79&4.67\\
    &LiDPM&0.446&0.532&0.440&34.09&19.45&6.27\\
    &ScoreLiDAR&0.406&--&0.425&--&--&--\\
    &LiNeXt&\underline{0.214}&\underline{0.494}&\underline{0.336}&\underline{41.07}&19.45&6.30\\
    \cmidrule{2-8}
    &LiDiff\textsuperscript{†}&0.376&0.573&0.416&32.43&22.99&13.40\\
    &ScoreLiDAR\textsuperscript{†}&0.342&--&0.399&--&--&--\\
    &LiDPM\textsuperscript{†}&0.376&0.542&0.403&36.59&\underline{25.76}&\underline{14.93}\\    &LiNeXt\textsuperscript{†}&\textbf{0.149}&\textbf{0.481}&\textbf{0.331}&\textbf{41.97}&\textbf{31.25}&\textbf{15.09}\\
    \midrule
    \multirow{10}[0]{*}{\begin{sideways}\rotatebox{0}{ KITTI-360}\end{sideways}}
    &LMSCNet&0.979&--&0.496&26.17&9.21&2.88\\
    &LODE&1.565&--&0.483&33.06&15.24&4.68\\
    &MID&0.637&--&0.476&33.05&21.32&11.30\\
    &LiDiff&0.564&--&0.459&33.23&17.55&4.88\\
    &ScoreLiDAR&0.472&--&0.444&--&--&--\\
    &LiNeXt&\underline{0.217}&\underline{0.508}&\underline{0.355}&\underline{36.85}&16.91&5.73\\
    \cmidrule{2-8}
    &LiDiff\textsuperscript{†}&0.517&--&0.446&33.43&\underline{22.04}&\underline{11.84}\\
    &ScoreLiDAR\textsuperscript{†}&0.452&--&0.437&--&--&--\\
    &LiNeXt\textsuperscript{†}&\textbf{0.149}&\textbf{0.499}&\textbf{0.339}&\textbf{41.88}&\textbf{29.34}&\textbf{13.90}\\
    \bottomrule
    \end{tabular}
    \caption{Comparison of various methods on the scene completion task on SemanticKITTI and KITTI-360. \textsuperscript{†} indicates additional refinement. Best results are highlighted in bold and second-best in underlined.}
    \label{tab:semantickitti_comparison}
\end{table*}

\section{Experiment}

\subsection{Experimental Settings}

\subsubsection{Dataset Preparation}
We train our model solely on the SemanticKITTI dataset~\cite{semantickitti}. Following LiDiff’s protocol~\cite{lidiff}, scans from sequences 00–10 are concatenated using the provided ego‑poses and filtered to remove all dynamic objects, resulting in dense static point clouds. Sequence 08 is reserved for validation. To assess cross‑dataset generalization, the pretrained model is evaluated without fine‑tuning on sequence 00 of the KITTI‑360 dataset~\cite{kitti360}.

\subsection{Evaluation Metrics}
We evaluate the completion performance using the Chamfer Distance~\cite{718487CD} (CD), Jensen–Shannon Divergence~\cite{1997JSD} (JSD) computed in both bird’s-eye view (BEV) and 3D space, as well as occupancy IoU at multiple voxel resolutions (0.5\,m, 0.2\,m, and 0.1\,m).

\section{Quantitative and Qualitative Comparisons} 

\subsection{Quantitative Comparisons}

LiNeXt consistently surpasses LiDiff on SemanticKITTI (Table~\ref{tab:semantickitti_comparison}): without refinement, Chamfer Distance decreases from 0.434 to 0.214, 3D JSD drops from 0.564 to 0.494, and BEV JSD from 0.444 to 0.336. The inference latency is 0.167s, which is 199.8× faster than that of LiDiff. The model size is 1.99M parameters, 16.4× smaller than LiDiff. With refinement, the CD further decreases to 0.149, the 3D JSD decreases to 0.481, and the BEV JSD decreases to 0.331. We evaluate the model trained on SemanticKITTI directly on KITTI‑360: LiNeXt\textsuperscript{†} retains its Chamfer Distance of 0.149 and achieves a marginal BEV‑JSD reduction of 0.008, whereas LiDiff\textsuperscript{†} degrades from a CD of 0.376 to 0.517, and its BEV‑JSD increases by 0.030.
This consistency across datasets highlights LiNeXt robustness and generalizability in various LiDAR scanning scenarios, demonstrating its ability to achieve high‑fidelity completion despite differing scene characteristics.
As shown in \cref{tab:performance_comparison}, LiNeXt attains an inference latency of 0.167\,s ($199.8\times$ faster than LiDiff) and 0.434\,s after refinement. Its model footprint is only $1.99 M$ parameters (2.10 M with refinement), corresponding to $16.4 \times$ and $25.9\times$ reductions compared to LiDiff and LiDiff\textsuperscript{†}, respectively. 
These results underscore the exceptional trade-off of LiNeXt among accuracy, speed and compactness.

\begin{table*}[ht]
    \centering
    \begin{tabular}{lcccccc}
    \toprule
    Method & CD↓ & JSD 3D↓ & JSD BEV↓ & Vox.IoU(0.5m)↑ & Vox.IoU(0.2m)↑ & Vox.IoU(0.1m)↑ \\
    \midrule
    LiNeXt & \textbf{0.214} & \textbf{0.494} & \textbf{0.336} & \textbf{41.07} & \textbf{19.45} & \underline{6.30} \\
    w/o DSR strategy & \underline{0.215} & 0.508 & 0.352 & \underline{40.00} & \underline{18.84} & \textbf{6.60} \\
    w/o MSSC module & 0.221 & \underline{0.502} & \underline{0.350} & 39.87 & 18.48 & 6.00 \\
    w/o CPA module & 0.227 & 0.504 & 0.353 & 39.36 & 18.65 & 5.84 \\

    \bottomrule
    \end{tabular}
    \caption{Ablation results on SemanticKITTI. Best results are highlighted in bold and second-best in underlined.}
    \label{tab:ablation_full}
\end{table*}

\begin{table}[h!]
    \centering
    \begin{tabular*}{\columnwidth}{@{\extracolsep{\fill}}lccc}
    \toprule
    Method&CD&Time(s)&Param(M)\\
    \midrule
    LiDiff&0.434&33.359&32.67\\
    LiDPM&0.446&15.288&32.67\\
    ScoreLiDAR&0.406&5.047&32.67\\
    LiNeXt&\underline{0.214}&\textbf{0.167}&\textbf{1.99}\\
    \midrule
    LiDiff\textsuperscript{†}&0.376&33.531&54.40\\
    LiDPM\textsuperscript{†}&0.377&15.453&54.40\\
    ScoreLiDAR\textsuperscript{†}&0.342&5.189&54.40\\
    LiNeXt\textsuperscript{†}&\textbf{0.149}&\underline{0.434}&\underline{2.10}\\
    \bottomrule
    \end{tabular*}
    \caption{Quantitative comparison of reconstruction accuracy, inference speed, and model size. \textsuperscript{†} denotes additional refinement. 
    The best and second-best results are highlighted in bold and underlined, respectively. 
    Inference speed is measured on a single NVIDIA RTX 3090 GPU.
    }
    \label{tab:performance_comparison}
\end{table}

\subsection{Qualitative Comparisons}
Figure~\ref{fig:Compare} compares scene completions on SemanticKITTI. Diffusion-based methods exhibit streak artifacts and density variations, whereas LiNeXt produces uniform, depth-consistent reconstructions. The refined LiNeXt\textsuperscript{†} further suppresses noise and fills occlusions, yielding artifact-free point clouds. These qualitative gains corroborate our quantitative metrics, confirming LiNeXt superior spatial uniformity and reconstruction fidelity.

\subsection{Ablation Study}

Our ablation study on SemanticKITTI (Table~\ref{tab:ablation_full}) isolates the contributions of LiNeXt core modules. First, removing the Distance‑Aware Selected Repeat (DSR) strategy markedly degrades all evaluation metrics except voxel IoU at 0.1m, highlighting DSR critical role in maintaining global shape coherence. Second, removing the Multi‑Scale Sparse Convolution (MSSC) module degrades all voxel IoU metrics (e.g., 39.87\% vs.\ 41.07\% at 0.5 m), underscoring MSSC’s role in recovering fine structural details. Finally, for the Cross‑Point Attention (CPA) ablation, we replace CPA with a standard cross‑attention mechanism only within the Noise‑to‑Coarse (N2C) Module’s hierarchical seed generation stage-full replacement would incur quadratic complexity and exceed our 24 GB memory budget. This partial substitution still incurs the largest performance drop (CD rises to 0.227, +6.1\%; IoU 0.5m falls to 39.36\%), confirming CPA’s critical role in hierarchical feature aggregation.

\section{Conclusion}

We have presented LiNeXt, a lightweight, non‑diffusion framework for 3D LiDAR scene completion. By introducing a Distance‑aware Selected Repeat strategy, a Noise‑to‑Coarse (N2C) Module, a Refine Module, a Cross‑Point Attention (CPA) mechanism, and a Multi‑Scale Sparse Convolution (MSSC) module, LiNeXt directly reconstructs complete point‑cloud scenes with high fidelity and efficiency. Extensive experiments on SemanticKITTI and KITTI‑360 demonstrate that LiNeXt achieves state‑of‑the‑art accuracy while delivering a 199.8$\times$ inference speedup and reducing model size to 6.1\% compared to LiDiff~\cite{lidiff}. These results underscore the practical suitability of LiNeXt for real-world  autonomous driving applications.

\section{Acknowledgements} 
This work was supported by and the National Natural Science Foundation of China (No.U25A20421, No.62202151, No.62202152) and the National Key Research and Development Program of China (No.2025YFB3003601).

\bibliography{Formatting-Instructions-LaTeX-2026}

\end{document}



\twocolumn[
    \thispagestyle{empty}
    \vspace*{1cm}
    \begin{center}
        {\LARGE \textbf{Supplementary Material for LiNeXt}}
    \end{center}
    \vspace{1.5cm}
]

\begin{table*}[t!]
    \centering
    \caption{Noise level ablation results on SemanticKITTI. Best results are highlighted in \textbf{bold} and second-best in \underline{underlined}.}
    \label{tab:ablation_noise}
    \begin{tabular}{lcccccc}
    \toprule
    Method & CD↓ & JSD 3D↓ & JSD BEV↓ & Vox.IoU(0.5m)↑ & Vox.IoU(0.2m)↑ & Vox.IoU(0.1m)↑ \\
    \midrule
    LiNeXt ($\sigma=0$) & 0.264 & 0.570 & 0.435 & 25.75 & 7.02 & 1.71 \\
    LiNeXt ($\sigma=0.5$) & \underline{0.216} & \underline{0.496} & 0.347 & 40.64 & \underline{19.27} & \textbf{6.34} \\
    LiNeXt ($\sigma=1.0$) & \textbf{0.214} & \textbf{0.494} & \textbf{0.336} & \textbf{41.07} & \textbf{19.45} & \underline{6.30} \\
    LiNeXt ($\sigma=2.0$) & \underline{0.216} & \underline{0.496} & \underline{0.344} & 40.88 & 19.24 & 6.05 \\
    LiNeXt ($\sigma=3.0$) & 0.218 & 0.499 & 0.345 & \underline{41.05} & 19.00 & 5.85 \\
    \bottomrule
    \end{tabular}
\end{table*}

\section{Experiment protocol}
\subsection{Implementation Details}
We implemented LiNeXt in PyTorch and trained on two NVIDIA RTX 3090 GPUs. For the Noise-to-Coarse (N2C) stage, we inject isotropic Gaussian noise \(\mathcal{N}(0,1)\) into the input coordinates. We train for 10 epochs with a batch size of 2 under the Adam optimizer, using an initial learning rate of $2\times10^{-4}$ and a weight decay of $1\times10^{-4}$. Subsequently, the Refine Module is configured for detail enhancement with an up‑sampling factor of 6 and trained for 5 epochs using the same batch size and decay schedule, but with an elevated initial learning rate of $5\times10^{-4}$. Each stage requires approximately 1 day of training. At test time, we evaluated on a single NVIDIA RTX 3090 GPU with a batch size of 1.

\subsection{Runtime Evaluation}
Runtime evaluations were performed on Sequence 08 of the SemanticKITTI\cite{semantickitti} dataset. To guarantee consistency, all non-critical services on the evaluation server were disabled. Each model was first primed with five input frames and then inference time was recorded over the following one hundred frames, with the average time across these frames reported as the model inference latency.

\subsection{Evaluation Metrics}
\label{sec:eval_metrics}

We employ a comprehensive set of metrics to quantitatively evaluate scene completion performance:

\textbf{Chamfer Distance (CD)}~\cite{718487CD}:
Measures bidirectional point‑set similarity between the predicted point cloud \(\mathcal{P}\) and the ground‑truth point cloud \(\widehat{\mathcal{P}}\):
\begin{equation}
    \mathrm{CD}(\mathcal{P}, \widehat{\mathcal{P}})
    = \frac{1}{|\mathcal{P}|}
    \sum_{x \in \mathcal{P}}
    \min_{y \in \widehat{\mathcal{P}}}
    \|x - y\|_2^2
    \;+\;
    \frac{1}{|\widehat{\mathcal{P}}|}
    \sum_{y \in \widehat{\mathcal{P}}}
    \min_{x \in \mathcal{P}}
    \|y - x\|_2^2.
\end{equation}
CD captures local geometric fidelity by averaging squared nearest‑neighbor distances in both directions; lower values indicate more accurate reconstructions.

\textbf{Jensen–Shannon Divergence (JSD)}~\cite{1997JSD}:  
To compute JSD, the predicted and ground‑truth point clouds are first voxelized into discrete distributions \(\mathcal{D}_{\mathrm{pred}}\) and \(\mathcal{D}_{\mathrm{gt}}\) on a 0.5 m grid. The divergence is then evaluated in both bird’s‑eye view (BEV) and full 3D space:
\begin{equation}
    \mathrm{JSD}(\mathcal{D}_{\mathrm{pred}}\parallel \mathcal{D}_{\mathrm{gt}})
    = \tfrac{1}{2}\,\mathrm{KL}\bigl(\mathcal{D}_{\mathrm{pred}}\parallel M\bigr)
    + \tfrac{1}{2}\,\mathrm{KL}\bigl(\mathcal{D}_{\mathrm{gt}}\parallel M\bigr),
\end{equation}
where \(\mathrm{KL}(\cdot\parallel\cdot)\) denotes the Kullback–Leibler divergence. and the mixture distribution \(M\) is defined as \(M = \tfrac{1}{2}\bigl(\mathcal{D}_{\mathrm{pred}} + \mathcal{D}_{\mathrm{gt}}\bigr)\).

\textbf{Multi-resolution Occupancy IoU}:  
To evaluate volumetric overlap, the predicted and ground-truth point clouds are first voxelized at resolutions \(r \in \{0.5\,\mathrm{m},\,0.2\,\mathrm{m},\,0.1\,\mathrm{m}\}\). At each resolution \(r\), the Intersection over Union is computed as
\[
    \mathrm{IoU}_r 
    = \frac{\lvert \mathcal{V}_{\mathrm{pred}}^r \cap \mathcal{V}_{\mathrm{gt}}^r\rvert}
           {\lvert \mathcal{V}_{\mathrm{pred}}^r \cup \mathcal{V}_{\mathrm{gt}}^r\rvert},
\]
where \(\mathcal{V}_{\mathrm{pred}}^r\) and \(\mathcal{V}_{\mathrm{gt}}^r\) denote the sets of occupied voxels at resolution \(r\). Reporting \(\mathrm{IoU}_r\) across these resolutions highlights reconstruction fidelity at multiple spatial granularities.

\section{Serial-Segment Max-Pooling (SSMP)}

In point-based feature aggregation, k-nearest neighbor (KNN) retrieval produces an unordered set of local neighbors, which precludes straightforward segment-wise max-pooling and dimensionality reduction. To impose a consistent ordering and enable structured downsampling, each neighborhood is first serialized into a one-dimensional sequence using spatial ordering. Specifically, this serialization is applied to the following:
\begin{itemize}
  \item the raw input point cloud \(P_{input}\), 
  \item the noisy point cloud \(P_{noise}\), 
  \item each down-sampled point set produced by FPS in the N2C module’s feature-extraction stage, 
  \item the final coarse output point cloud \(P_{coarse}\) from the N2C module. 
\end{itemize}
To preserve spatial locality, we randomly choose either Z-order~\cite{1966z-order} or Hilbert-order~\cite{hilbert1935stetige} for the serialization.

\begin{figure}[t]
\centering
\includegraphics[width=1.00\columnwidth]{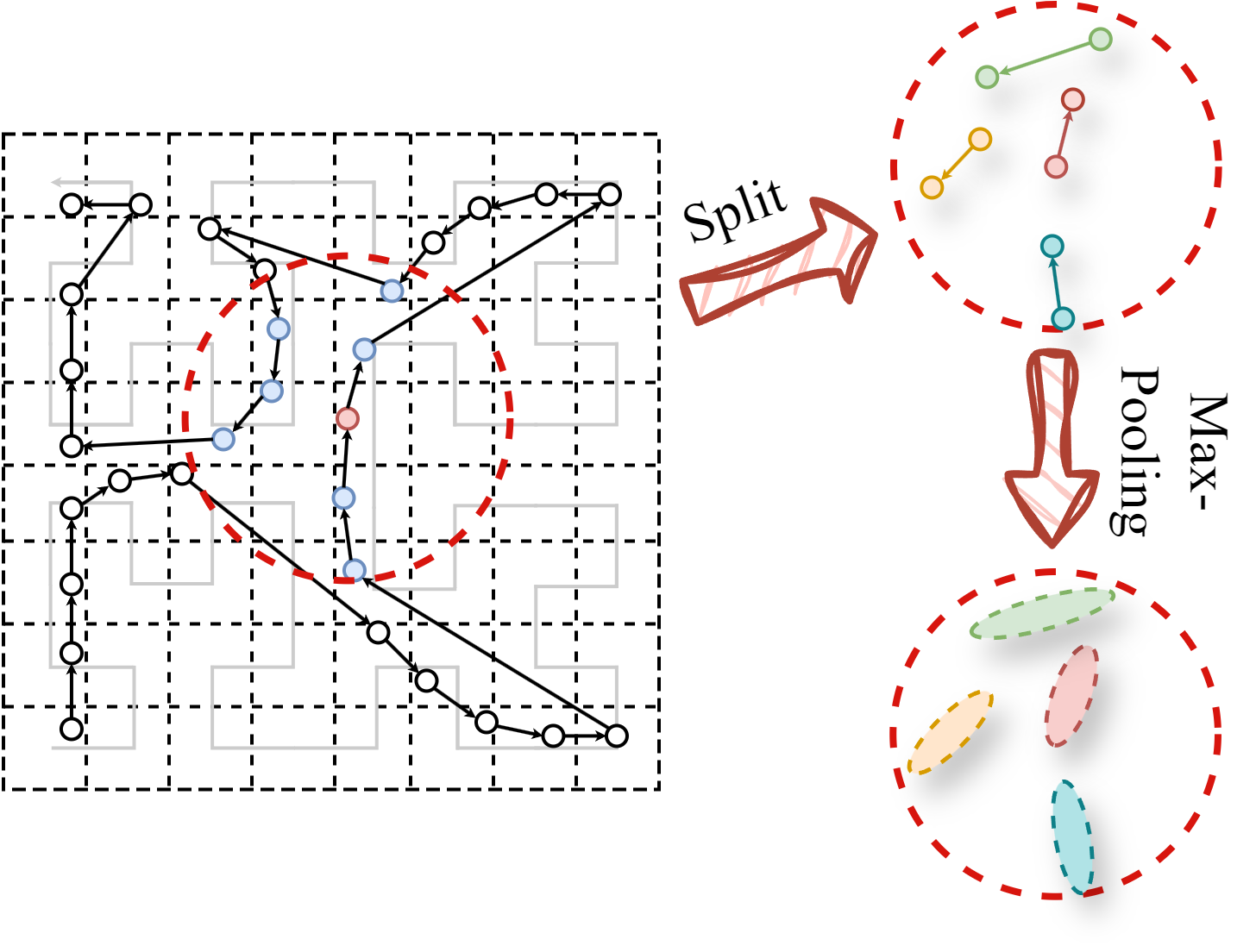} 
\caption{Illustration of the Serial-Segment Max-Pooling (SSMP) procedure.}
\label{fig:SSMP}
\end{figure}

As illustrated in the 2D demonstration in Figure~\ref{fig:SSMP}, SSMP initially partitions the serialized, ordered neighborhood of each point into \(\hat{K}\) equal‑length segments, and then applies max‑pooling within each segment to produce a compressed, abstract feature representation.

Specifically, SSMP operates on a point-wise relational tensor of shape \(\mathbb{R}^{N\times C\times K}\), where \(N\) is the number of points, \(C\) the feature dimension, and \(K\) the number of neighbors per point after serialization. We partition the third dimension into \(\hat{K}\) contiguous segments of size \(K/\hat{K}\), yielding a tensor of shape \(\mathbb{R}^{N\times C\times \hat{K}\times (K/\hat{K})}\). Each segment preserves the sequence-aware ordering that is induced by the chosen space-filling curve.

We then apply max-pooling along the innermost dimension of length \(K/\hat{K}\), compressing each segment into a single representative feature of maximal activation. The result is a condensed relational tensor of shape \(\mathbb{R}^{N\times C\times \hat{K}}\), which captures the most salient local patterns within each serial segment. By segmenting the pooling in this way, SSMP achieves fine-grained spatial abstraction while retaining the global sequence structure of the neighborhood.

\section{Noise Level Ablation}
We perform an ablation study on the SemanticKITTI dataset to assess the robustness of the model against the isotropic Gaussian noise $\mathcal{N}(0, \sigma^2)$ injected into the input point coordinates. 

Table~\ref{tab:ablation_noise} quantifies performance sensitivity to varying noise scales $\sigma$. The optimal setting, \(\sigma=1.0\), yields the best results across most metrics, achieving a Chamfer Distance of 0.214, 3D JSD of 0.494, BEV JSD of 0.336, and voxel IoU of 41.07\% at 0.5m and 19.45\% at 0.2m. Performance deteriorates when \(\sigma\) deviates from this optimum: zero noise (\(\sigma=0\)) increases CD by 26.1\%, while excessive noise (\(\sigma\ge2.0\)) impairs fine-detail reconstruction (voxel IoU of 6.05–5.85\% at 0.1m versus 6.30\% at \(\sigma=1.0\)).

These results demonstrate that balanced noise injection ($\sigma=1.0$) combined with our proposed modules yields optimal noise robustness while preserving structural fidelity across scales.

\section{Additional Qualitative Results}
In this section, we provide additional qualitative comparisons to highlight LiNeXt’s superior scene completion performance on the SemanticKITTI~\cite{semantickitti} and KITTI‑360~\cite{kitti360} datasets. We compare LiNeXt with recent state‑of‑the‑art methods, including LiDiff\cite{lidiff} and ScoreLiDAR\cite{scorelidar}, and observe that our method consistently produces more complete and geometrically accurate reconstructions across a variety of complex scenes.

\subsection{SemanticKITTI}

\begin{figure*}[t!]
  \centering
  \includegraphics[width=\textwidth]{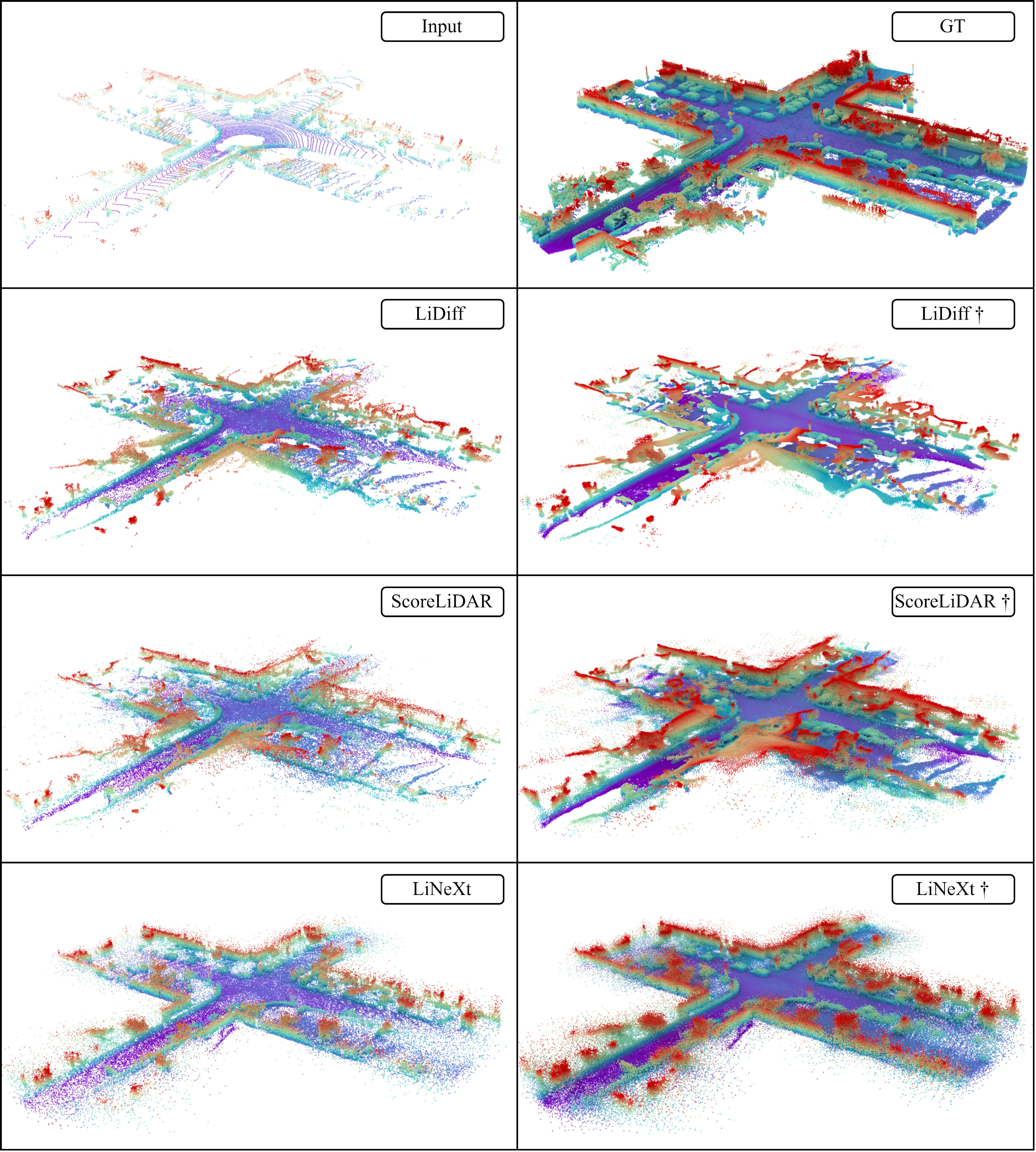}
  \caption{Visualization comparison of our method against LiDiff and ScoreLiDAR on the SemanticKITTI dataset. \textsuperscript{†} indicates additional refinement.}
  \label{fig:Compare_3}
\end{figure*}

\begin{figure*}[t!]
  \centering
  \includegraphics[width=\textwidth]{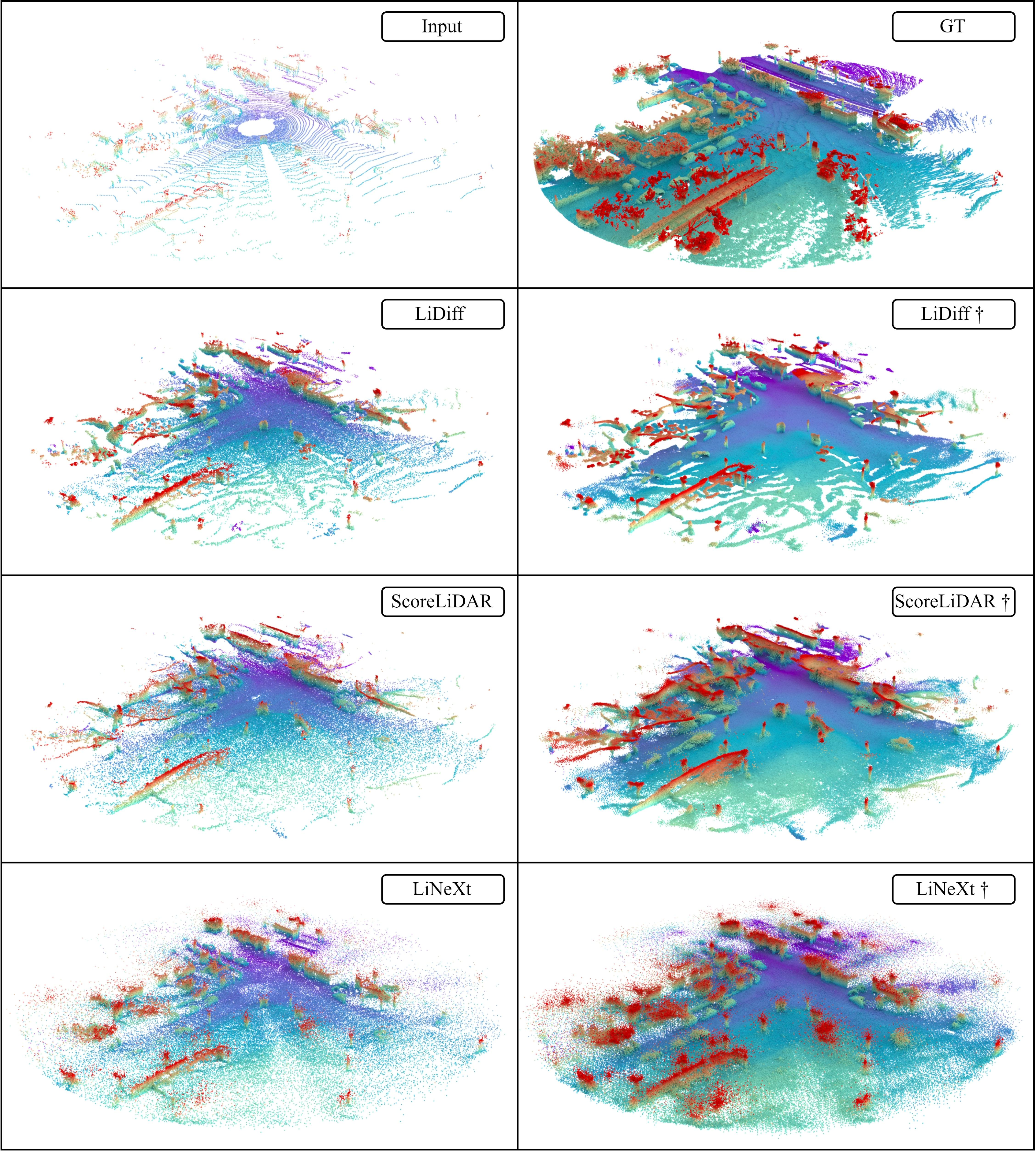}
  \caption{Visualization comparison of our method against LiDiff and ScoreLiDAR on the SemanticKITTI dataset. \textsuperscript{†} indicates additional refinement.}
  \label{fig:Compare_4}
\end{figure*}

\begin{figure*}[t!]
  \centering
  \includegraphics[width=\textwidth]{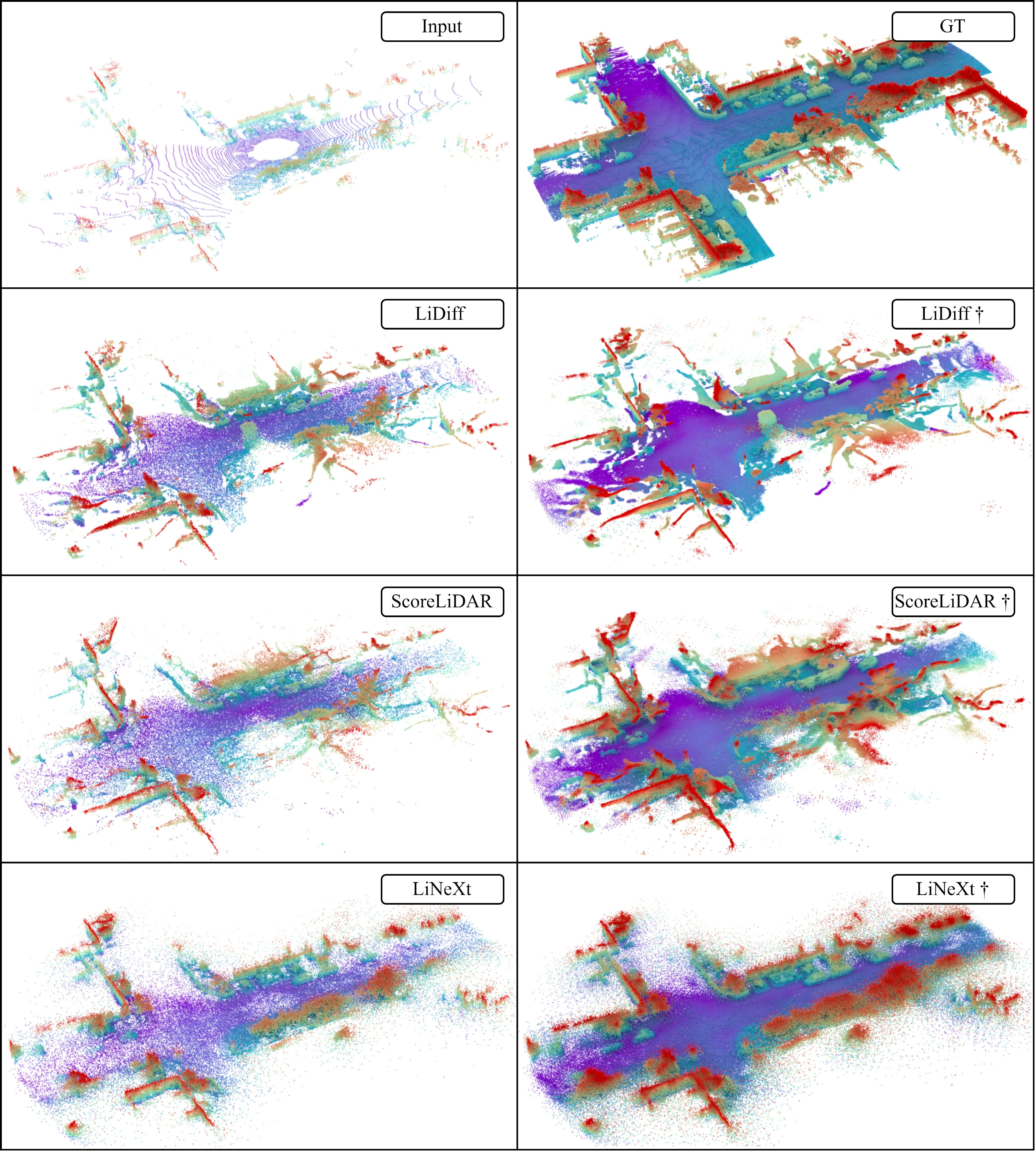}
  \caption{Visualization comparison of our method against LiDiff and ScoreLiDAR on the SemanticKITTI dataset. \textsuperscript{†} indicates additional refinement.}
  \label{fig:Compare_2}
\end{figure*}

Figures~\ref{fig:Compare_3}, ~\ref{fig:Compare_4}, and~\ref{fig:Compare_2} illustrate representative scenes from SemanticKITTI. LiNeXt consistently yields more accurate reconstructions of vehicles, street furniture, and terrain, exhibiting smooth surfaces and precise contours. By contrast, LiDiff and ScoreLiDAR exhibit stripe-like bands of missing or oversmoothed points, creating unrealistic artifacts that misalign with the true geometry. For example, in Figure~\ref{fig:Compare_3}, LiNeXt faithfully reproduces the vehicle’s shape and outline, whereas LiDiff suffers from partial point cloud loss and ScoreLiDAR yields an indistinct silhouette. Additionally, the diffusion methods introduce stripe-shaped point clusters on adjacent walls, deviating from their planar surfaces.

\subsection{KITTI-360}

\begin{figure*}[t!]
  \centering
  \includegraphics[width=\textwidth]{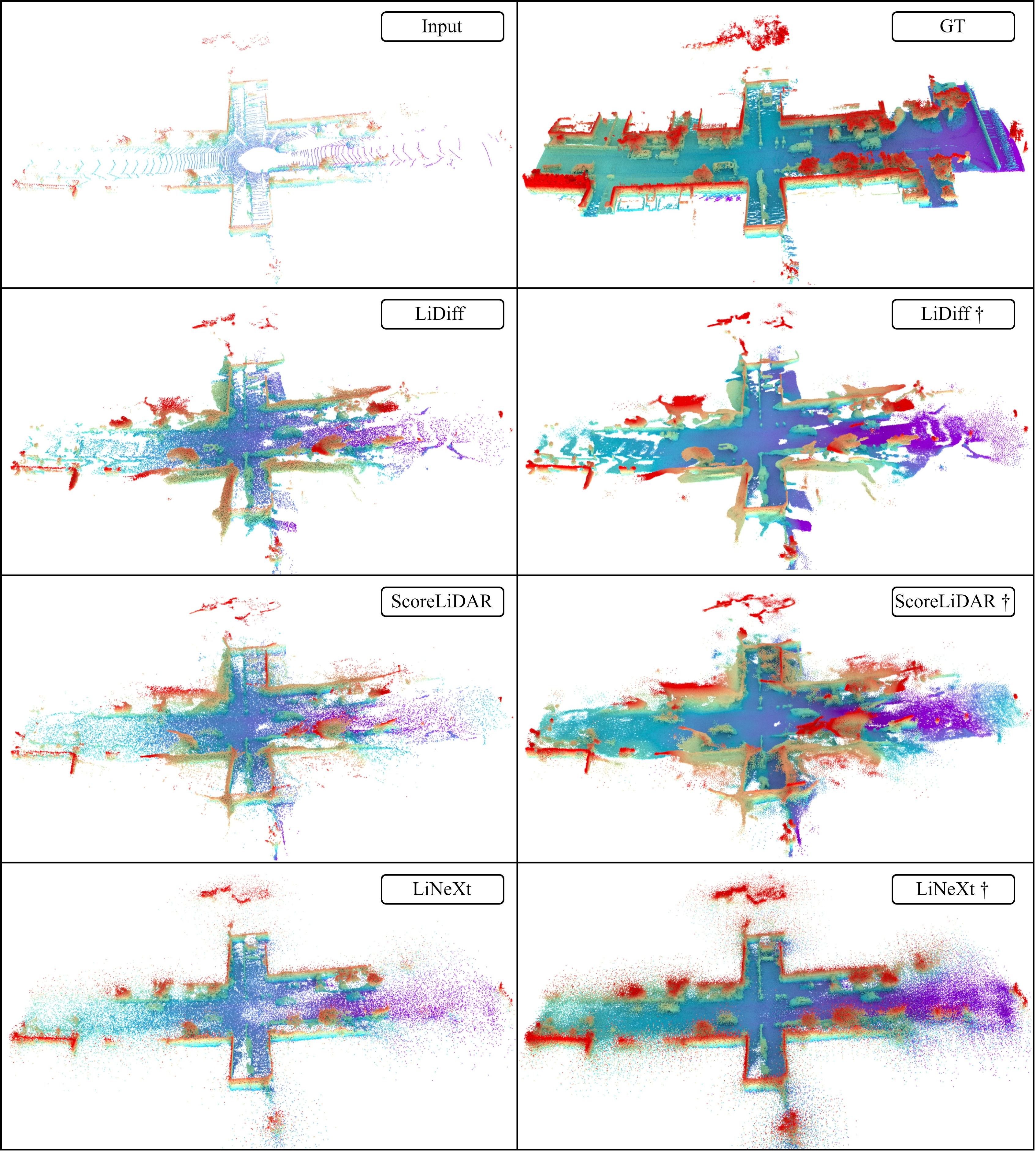}
  \caption{Visualization comparison of our method against LiDiff and ScoreLiDAR on the KITTI-360 dataset. \textsuperscript{†} indicates additional refinement.}
  \label{fig:KITTI-360_2}
\end{figure*}

\begin{figure*}[t!]
  \centering
  \includegraphics[width=\textwidth]{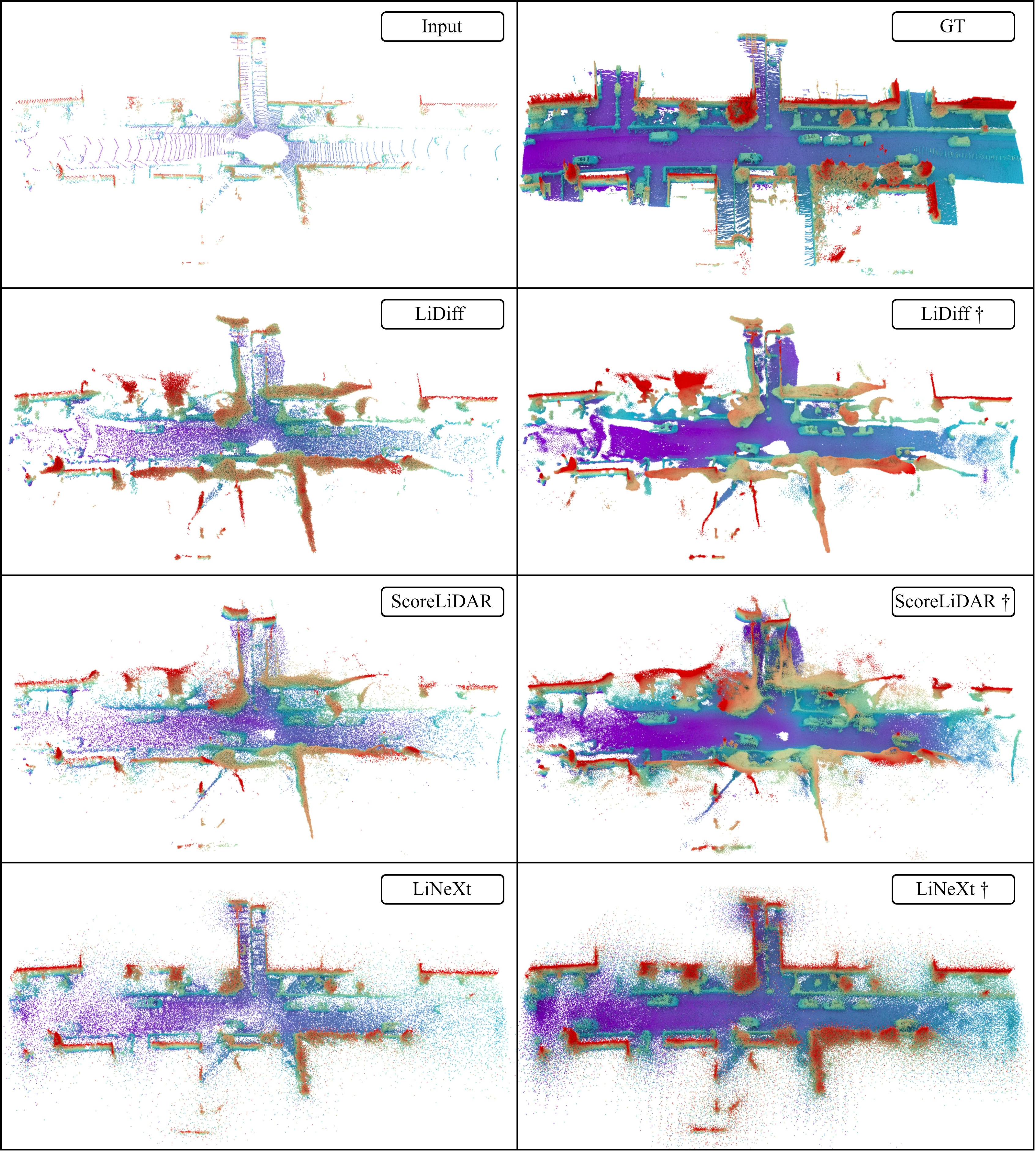}
  \caption{Visualization comparison of our method against LiDiff and ScoreLiDAR on the KITTI-360 dataset. \textsuperscript{†} indicates additional refinement.}
  \label{fig:KITTI-360_3}
\end{figure*}

We applied the model, trained solely on SemanticKITTI, to the KITTI-360 dataset without fine-tuning. Figures~\ref{fig:KITTI-360_2} and~\ref{fig:KITTI-360_3} present these cross-dataset results. LiNeXt maintains structural regularity by preserving planar facades and continuous road surfaces, effectively avoiding the stripe artifacts observed in LiDiff and ScoreLiDAR. In Figure~\ref{fig:KITTI-360_2}, LiNeXt accurately reconstructs vehicles and trees, delivering a globally coherent scene.

This zero-shot transfer highlights LiNeXt’s strong generalizability across datasets, eliminating the need for dataset-specific retraining and enabling rapid deployment in diverse environments. Overall, LiNeXt’s lightweight, non‑diffusion architecture is well suited for real-world LiDAR scene completion, combining high fidelity with artifact-free reconstructions.

\section{Failure Example}

\begin{figure*}[t!]
  \centering
  \includegraphics[width=\textwidth]{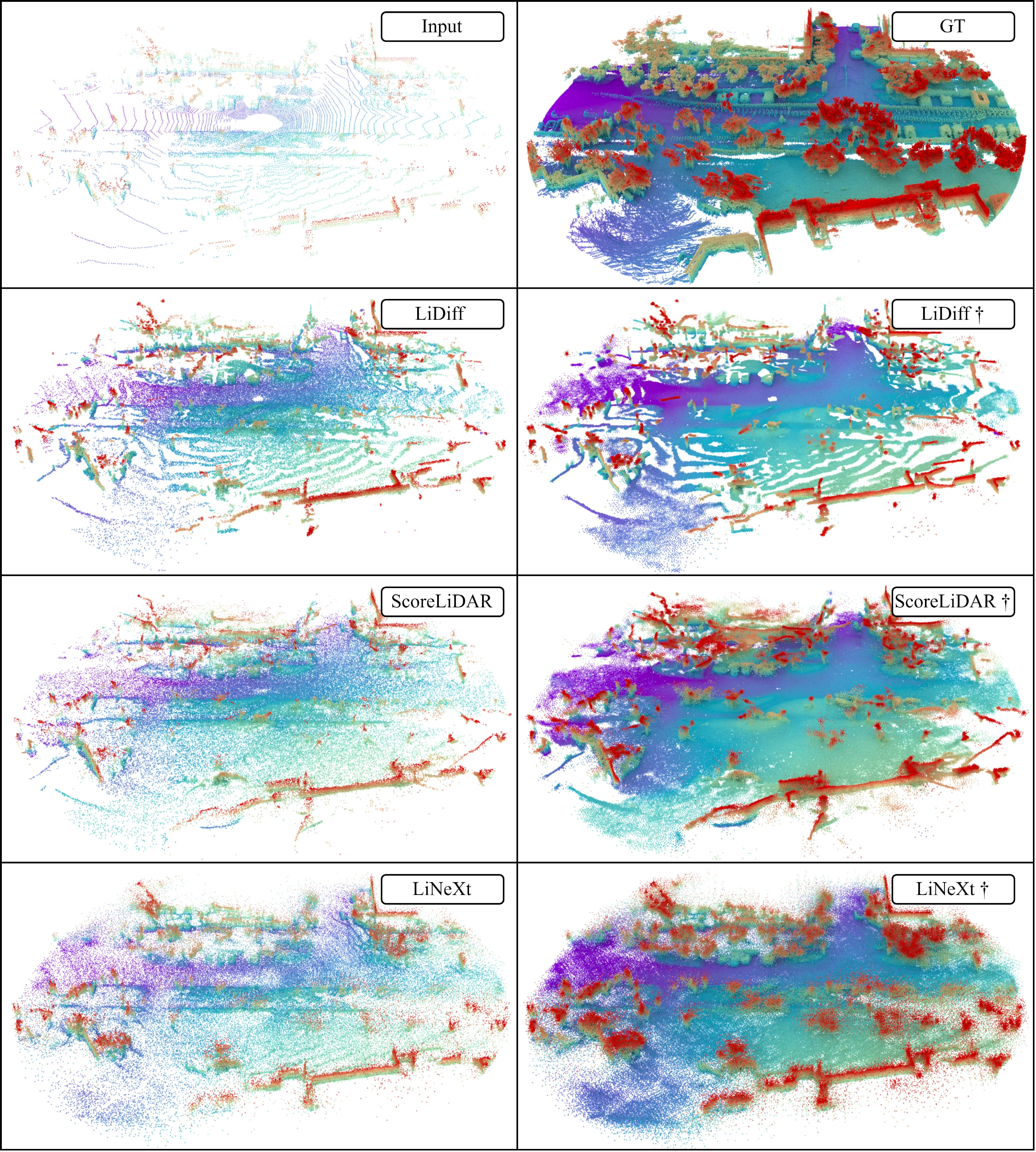}
  \caption{Visualization of failure cases produced by our method, LiDiff and ScoreLiDAR on the KITTI-360 dataset. \textsuperscript{†} indicates additional refinement.}
  \label{fig:KITTI-360_1}
\end{figure*}

As shown in Figure~\ref{fig:KITTI-360_1}, in scenes with numerous elements and severe occlusions, LiNeXt occasionally produces floating-point artifacts, where reconstructed points detach from the main object bodies and hover in mid‑air. By comparison, LiDiff exhibits substantial geometry loss and blurred or incomplete object outlines, and ScoreLiDAR generates irregular point clusters and over‑smoothed surfaces that distort fine details. These failure modes highlight the persistent challenges of reconstructing heavily occluded regions and densely cluttered environments.

\section{Future Work}

Future research will pursue several complementary directions to further advance LiDAR scene completion:

\begin{enumerate}
    \item \textbf{Uniform Point Cloud Generation:} Investigate techniques for generating synthetic point clouds with spatially uniform sampling and complete structural coverage. This includes leveraging adversarial regularization and occupancy priors to mitigate irregular density distributions and fill missing regions in large-scale outdoor scans.
    
    \item \textbf{Multimodal Fusion:} Integrate complementary data modalities such as RGB imagery, depth maps, and inertial measurements within a unified completion framework. Cross-modal attention mechanisms and learned projection consistency constraints will be explored to enforce alignment between LiDAR and image domains.
    
    \item \textbf{Semantic Scene Completion:} Extend LiNeXt to perform joint geometry–semantics inference by incorporating point-wise classification heads and semantic consistency losses, enabling simultaneous prediction of object labels and completion of missing surfaces for downstream perception tasks.
    
    \item \textbf{Temporal Consistency:} Enforce coherence across sequential LiDAR scans by incorporating temporal smoothness constraints, ensuring consistent and stable reconstructions in dynamic environments.
    
    \item \textbf{Real-Time Deployment:} Optimize network architectures and inference pipelines to meet real-time processing requirements through methods such as model pruning, quantization, and depthwise convolutions, reducing computational overhead without compromising reconstruction quality.
\end{enumerate}

\bibliography{LiNeXt_Supp}